\documentclass{article}

    \PassOptionsToPackage{numbers, sort&compress}{natbib}
\usepackage[preprint]{neurips_2026}

\usepackage[utf8]{inputenc} 
\usepackage[T1]{fontenc}    
\usepackage{hyperref}       
\usepackage{url}            
\usepackage{booktabs}       
\usepackage{amsfonts}       
\usepackage{nicefrac}       
\usepackage{microtype}      
\usepackage{xcolor}         
\usepackage{graphicx} 
\usepackage{array}
\usepackage{subcaption}
\usepackage{amsmath} 
\usepackage{titlesec}
\usepackage{enumitem} 
\titlespacing*{\section}{0pt}{2pt plus 1pt minus 1pt}{1pt}
\titlespacing*{\subsection}{0pt}{2pt plus 1pt minus 1pt}{1pt}
\titlespacing*{\subsubsection}{0pt}{2pt plus 1pt minus 1pt}{1pt}

\title{Evo-Depth: A Lightweight Depth-Enhanced Vision-Language-Action Model}

  \author{%
  \mdseries
  Tao Lin\textsuperscript{1,*} \quad                                  
  Yuxin Du\textsuperscript{1,*} \quad                                  
  Jiting Liu\textsuperscript{1,*} \quad
  Nuobei Zhu\textsuperscript{1} \quad
  Yunhe Li\textsuperscript{1} \quad
  Yuqian Fu\textsuperscript{2} \\
  Yinxinyu Chen\textsuperscript{1} \quad
  Hongyi Cai\textsuperscript{1} \quad
  Zewei Ye\textsuperscript{1} \quad
  Bing Cheng\textsuperscript{1,4} \quad
  Kai Ye\textsuperscript{1} \\
  Yiran Mao\textsuperscript{1} \quad
  Yilei Zhong\textsuperscript{1} \quad
  MingKang Dong\textsuperscript{1} \quad
  Junchi Yan\textsuperscript{1} \quad
  Gen Li\textsuperscript{3,\dag} \quad
  Bo Zhao\textsuperscript{1,4,\dag} \\[1em]
  \textsuperscript{1}School of AI, Shanghai Jiao Tong University \\
  \textsuperscript{2}King Abdullah University of Science and Technology \\
  \textsuperscript{3}Nanyang Technological University 
  \textsuperscript{4}SJTU-Quic Robot Joint Lab \\[0.5em]
  taolin200108@gmail.com, bo.zhao@sjtu.edu.cn \\
  \href{https://github.com/MINT-SJTU/Evo-Depth}{https://github.com/MINT-SJTU/Evo-Depth} \\[0.5em]
  \textsuperscript{*}Equal contribution \quad
  \textsuperscript{\dag}Corresponding authors
  }

\begin{document}

\maketitle

\begin{abstract}
Vision-Language-Action (VLA) models have emerged as a promising paradigm for robotic manipulation by unifying perception, language grounding, and action generation. However, they often struggle in scenarios requiring precise spatial understanding, as current VLA models primarily rely on 2D visual representations that lack depth information and detailed spatial relationships. 
While recent approaches incorporate explicit 3D inputs such as depth maps or point clouds to address this issue, they often increase system complexity, require additional sensors, and remain vulnerable to sensing noise and reconstruction errors. 
Another line of work explores implicit 3D-aware spatial modeling directly from RGB observations without extra sensors, but it often relies on large geometry foundation models, resulting in higher training and deployment costs.
To address these challenges, we propose \textbf{Evo-Depth}, a lightweight depth-enhanced VLA framework that enhances spatially grounded manipulation without relying on additional sensing hardware or compromising deployment efficiency.
Evo-Depth employs a lightweight \textit{Implicit Depth Encoding Module} (IDEM) to extract compact depth features from multi-view RGB images. These features are incorporated into vision-language representations through a \textit{Spatial Enhancement Module} (SEM) via depth-aware modulation, enabling efficient spatial-semantic enhancement. A \textit{Progressive Alignment Training} strategy is further introduced to align the resulting depth-enhanced representations with downstream action learning.
Extensive experiments in simulation and real-world settings demonstrate the effectiveness of Evo-Depth. 
With only \textbf{0.9B} parameters, Evo-Depth achieves superior performance across four simulation benchmarks.
In real-world experiments, Evo-Depth attains the highest average success rate while also exhibiting the smallest model size, lowest GPU memory usage, and highest inference frequency among compared methods. These results demonstrate that lightweight implicit depth enhancement is an effective and practical solution for spatially grounded robotic manipulation. We will release code and models to facilitate future research on lightweight depth-enhanced VLA models.

\end{abstract}

\section{Introduction}
Recent advances in embodied intelligence have increasingly emphasized the     
importance of unifying perception, language, and control for general-purpose
robotic manipulation, giving rise to Vision-Language-Action (VLA)          
models~\cite{xie2025dexbotic,team2025gigabrain,yu2026twinbrainvla,lian2026langforce,luo2026coral,bjorck2025gr00t}. By
grounding natural language instructions in visual observations and directly
generating actions, VLA models provide a promising framework for scalable
robotic learning. Despite this progress, their performance still degrades
noticeably in manipulation scenarios that require precise spatial
understanding.

A major limitation of current VLA models~\cite{kim2024openvla,intelligence2025pi,lin2025evo1} is that they rely primarily on 2D visual representations, which provide limited depth information and spatial relationships for precise manipulation. As a result, their performance often drops on tasks requiring accurate localization, fine-grained placement, and spatially consistent manipulation. 
To address this issue, recent studies have introduced explicit 3D information, such as depth maps or point clouds, into robotic policy learning~\cite{qu2025spatialvla,li2026pointvla,zhen20243d}. However, reliance on additional sensing hardware or explicit geometry-processing pipelines often increases system complexity, making these methods sensitive to sensing noise and reconstruction errors, which in turn limits their practicality and scalability in real-world deployment. 
Another line of work explores implicit 3D-aware spatial modeling from RGB observations
alone~\cite{lin2025evo0,li2025spatial,abouzeid2025geoaware,yu20263d,qian2025gp3}, avoiding explicit 3D sensing and providing a more scalable alternative.
Nevertheless, these methods often introduce large-scale geometry foundation models, such as VGGT~\cite{wang2025vggt}. This increases model size and raises training and deployment costs, making them less suitable for real-time robotic deployment.
At the same time, many existing fusion designs remain relatively heavy, leaving lightweight yet effective spatial-semantic fusion underexplored. In addition, how to effectively align such spatial enhancement with downstream action learning during training also remains an open challenge.

\begin{figure}[t]
  \centering
  \includegraphics[width=\linewidth]{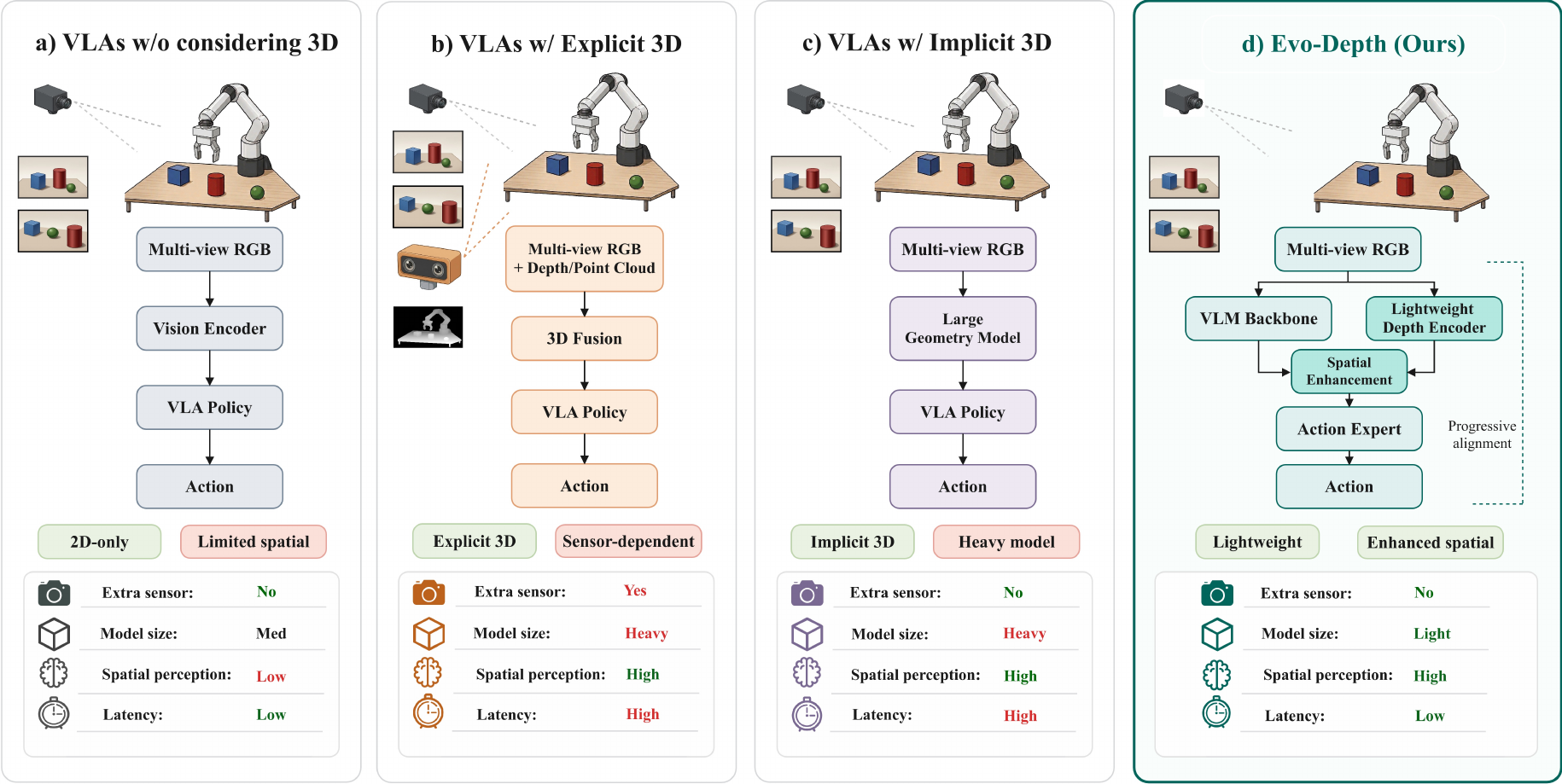}
  \caption{
  \textbf{Comparison of Evo-Depth with three representative VLA paradigms.}   
  (a) 2D-only VLAs are efficient but lack spatial perception.                   
  (b) Explicit 3D VLAs offer strong spatial understanding at the cost of extra  
  sensors and high latency.                  
  (c) Implicit 3D VLAs remove sensor dependency but rely on heavy geometry   
  models.
    (d) Evo-Depth improves spatial perception with low latency by using the lightweight IDEM module to extract depth-aware features and the SEM module for efficient spatial enhancement.
  }

  \label{fig:teaser}
\end{figure}

To address these challenges, we propose \textbf{Evo-Depth}, a lightweight depth-enhanced VLA framework for improving spatially grounded manipulation while preserving deployment efficiency, as illustrated in Fig.~\ref{fig:teaser}. 
Evo-Depth adopts a lightweight implicit depth enhancement strategy to strengthen spatial perception without relying on explicit 3D inputs or additional sensing hardware. 
It first extracts spatial information from multi-view RGB observations through a lightweight \textit{Implicit Depth Encoding Module}
(IDEM), which produces compact latent depth features that capture relative spatial relationships and object layout without constructing heavy geometry-rich scene representations.
However, such information alone is insufficient to support downstream decision making, as it needs to be effectively incorporated into vision-language representations. To this end, we introduce a \textit{Spatial Enhancement Module} (SEM), which uses implicit depth features as lightweight modulation signals to enhance visual features, rather than relying on heavy feature fusion, thereby preserving the original vision-language representations while injecting complementary spatial cues.
In addition, to better support end-to-end optimization for depth-enhanced policy learning, we propose a \textit{Progressive Alignment Training} strategy that gradually aligns spatial representations with downstream action learning. Together, these designs enable Evo-Depth to improve spatial understanding and manipulation performance without relying on explicit 3D inputs or additional sensing hardware, while maintaining a lightweight and deployable VLA framework for effective cross-modal fusion and end-to-end policy learning.

Extensive experiments in both simulation and real-world settings demonstrate that Evo-Depth achieves superior performance while maintaining low memory usage and high inference frequency. Notably, with only 0.9B parameters, Evo-Depth achieves 84.4\% on Meta-World, 41.1\% on VLA-Arena, 95.4\% on LIBERO, and 69.6\% on LIBERO-Plus on simulation benchmarks. 
In real-world experiments, Evo-Depth achieves the highest average success rate of 90\%, with the smallest model size, lowest GPU memory usage (3.2 GB), and highest inference frequency (12.3 Hz) among all compared methods. Moreover, under four disturbance settings in real-world generalization experiments, Evo-Depth consistently outperforms the baseline model, demonstrating improved robustness to unseen visual and spatial variations.

Our contributions are summarized as follows:
\begin{enumerate}[itemsep=2pt, topsep=2pt, parsep=0pt, partopsep=0pt]
\item We propose \textbf{Evo-Depth}, a lightweight and deployment-efficient depth-enhanced VLA framework that incorporates implicit depth information to improve spatially grounded manipulation without relying on explicit 3D inputs or additional sensing hardware.
  
\item We introduce a \textit{Spatial Enhancement Module} to support lightweight spatial-semantic fusion, together with a \textit{Progressive Alignment Training} strategy that progressively aligns depth-enhanced representations with downstream action learning.
  
    \item Extensive experiments in both simulation and real-world settings demonstrate the effectiveness of Evo-Depth. With only 0.9B parameters, Evo-Depth achieves strong performance across simulation benchmarks and attains the highest average success rate in real-world experiments, while maintaining low GPU memory usage and high inference frequency, making it well suited for real-time robotic deployment.
\end{enumerate}

\section{Related Work}

\subsection{Vision-Language-Action Models}
Vision-Language-Action (VLA) models~\cite{bu2025univla,cai2026xiaomi,luo2026being,luo2026simvla,shi2026saivla,wang2026vla,zheng2025x,zhong2026acot} provide a unified framework that grounds language instructions in visual observations and directly generates robot actions. Recent progress has shown that large-scale multimodal pretraining combined with end-to-end policy learning can substantially improve manipulation performance~\cite{wu2026pragmatic,black2410pi0,yang2026abot}.   Representative generalist VLA models include OpenVLA~\cite{kim2024openvla}, an open-source model pretrained on large-scale real-world robot demonstrations~\cite{o2024open}; GR00T~\cite{bjorck2025gr00t}, which combines a vision-language module with a diffusion transformer for real-time action generation; and $\pi^{*}_{0.6}$~\cite{intelligence2025pi}, which further improves VLA performance through real-world reinforcement learning.
To improve deployment efficiency, recent work has also explored lightweight VLA models. TinyVLA~\cite{wen2025tinyvla} improves inference speed and data efficiency with a compact architecture, SmolVLA~\cite{shukor2025smolvla} combines a lightweight vision-language backbone with a flow-matching action expert, and Evo-1~\cite{lin2025evo1} further reduces computation while preserving semantic alignment. Despite these advances, most VLA models still
rely on 2D visual observations, which limits their effectiveness in tasks
requiring precise spatial understanding and highlights the need for 3D-aware or depth-informed representations.

\subsection{Vision-Language-Action Models with 3D Perception}

To overcome the spatial limitations of 2D-based policies, recent studies have incorporated explicit 3D information into robotic policy
learning~\cite{tan2026masked,deng2025stereovla,li2025qdepth,abouzeid2025geoaware,fan2026any3d,yang2026robo3r}. Representative examples include
3D-VLA~\cite{zhen20243d}, which unifies 3D perception, reasoning, and action in a generative model; SpatialVLA~\cite{qu2025spatialvla}, which introduces 3D spatial context through Ego3D Position Encoding and structured actions; and PointVLA~\cite{li2026pointvla}, which injects point cloud features into pretrained VLA models. These methods improve spatial understanding through richer geometric representations, but their reliance on explicit 3D inputs often requires additional sensing hardware and complex processing pipelines, increasing system complexity and sensitivity to sensing noise and reconstruction errors.
Recent studies have also explored implicit 3D-aware spatial modeling from RGB observations alone~\cite{lin2025evo0,li2025spatial,rao2026augvla,peng2025omnivggt}. Although these methods avoid explicit 3D sensing pipelines, they are generally built upon large-scale geometry foundation models~\cite{wang2025vggt}, resulting in larger model size and higher training and deployment costs. Together, these limitations motivate a lightweight implicit depth approach that enhances VLA models with depth-aware spatial information from RGB observations while preserving efficiency.

\section{Method}

\subsection{Overview of Evo-Depth Architecture}

\begin{figure}[t]
    \centering
    \includegraphics[width=\linewidth]{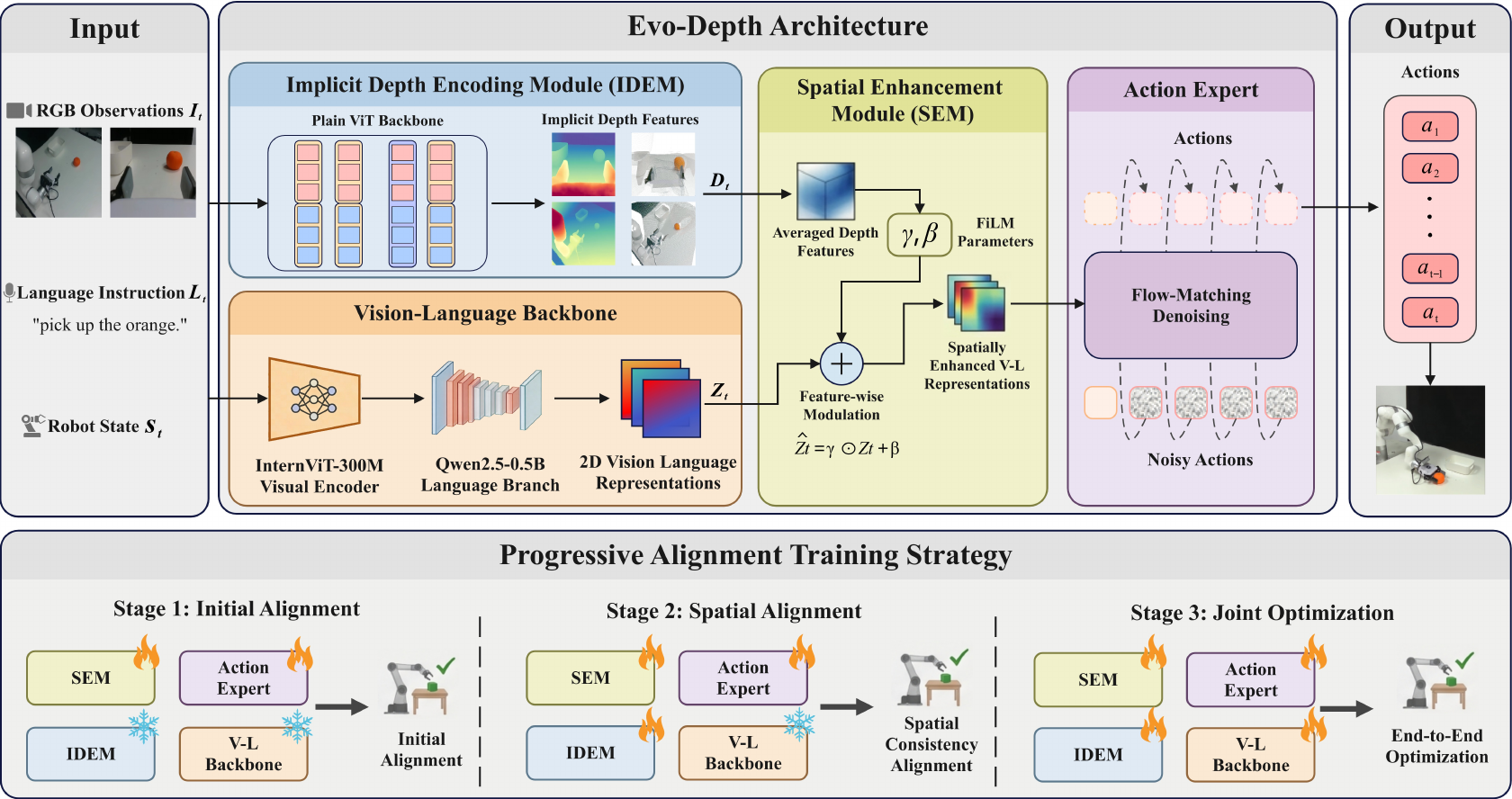}
    \caption{
    \textbf{Overview of Evo-Depth.} Given multi-view RGB observations, language instructions, and robot states, Evo-Depth first extracts implicit depth cues with the IDEM, then enhances vision-language representations through the SEM, and finally predicts actions with a flow-matching action expert. The bottom panel shows the progressive alignment training strategy, which improves cross-module coordination and facilitates effective fusion between implicit depth features and vision-language representations through staged optimization.
    }
    \label{fig:framework}
\end{figure}

Evo-Depth is a lightweight spatially enhanced Vision-Language-Action model that improves
spatially grounded manipulation by incorporating implicit depth cues into
policy learning, without requiring additional 3D sensors or explicit 3D
inputs. As shown in Fig.~\ref{fig:framework}, Evo-Depth consists of four
components: (1) \textit{Implicit Depth Encoding Module} (IDEM), (2)
\textit{Vision-Language Backbone}, (3) \textit{Spatial Enhancement Module}
(SEM), and (4) \textit{Action Expert}.

Given multi-view RGB observations $\{I_t^i\}_{i=1}^{N}$, a language
instruction $L_t$, and robot state $s_t$, Evo-Depth predicts the target action
trajectory as
\begin{equation}
a_t = f_{\mathrm{Evo\text{-}Depth}}\left(\{I_t^i\}_{i=1}^{N}, L_t, s_t;
\theta\right),
\end{equation}
where $\theta$ denotes all learnable parameters. Specifically, IDEM takes
$\{I_t^i\}_{i=1}^{N}$ as input and produces implicit depth features $D_t$. In
parallel, the vision-language backbone encodes $\{I_t^i\}_{i=1}^{N}$ and $L_t$
into 2D vision-language representations $Z_t$. SEM then takes $D_t$ and $Z_t$
as input and outputs spatially enhanced representations $\hat{Z}_t$. Finally,
the action expert predicts the target action trajectory $a_t$ conditioned on
$\hat{Z}_t$ and $s_t$.

\subsection{Model Design}

\noindent\textbf{Implicit Depth Encoding Module.}
The Implicit Depth Encoding Module (IDEM) aims to extract implicit depth features directly from multi-view RGB observations, without requiring additional 3D perception sensors or explicit 3D inputs. Formally, given multi-view RGB observations $\{I_t^i\}_{i=1}^{N}$, IDEM produces implicit depth features
\begin{equation}
D_t = f_{\mathrm{IDEM}}\left(\{I_t^i\}_{i=1}^{N}\right),
\end{equation}
where $D_t$ denotes the latent depth-aware representation used for subsequent spatial enhancement. 
Concretely, IDEM is built upon a single plain Vision Transformer backbone with only 0.13B parameters. Its weights are initialized from a multi-view depth encoder~\cite{lin2025depth} pretrained for joint any-view depth estimation. The pretrained encoder is learned with a minimal yet sufficient training objective for any-view depth estimation under large-scale, high-quality supervision, providing a strong initialization without relying on heavier architectural specialization.
To capture spatial relationships across different observations, IDEM is designed to progressively enable information exchange across views. 
Let $H^{(l)}$ denote the token representations at layer $l$. The
attention pattern is defined as
\begin{equation}
H^{(l+1)} =
\begin{cases}
\mathrm{Attn}_{\mathrm{within}}\!\left(H^{(l)}\right), & l < l_0, \\[4pt]
\mathrm{Attn}_{\mathrm{cross/within}}\!\left(H^{(l)}\right), & l \ge l_0,
\end{cases}
\end{equation}
where the first $l_0$ layers apply self-attention independently within each
view to preserve local image structure and establish stable view-specific
representations, while the remaining layers alternate between within-view and
cross-view attention to progressively enable information exchange across
observations.
Through this design, the encoder aggregates spatial cues from different views into a unified latent feature space.
The latent transformer embeddings are taken as the output of IDEM and serve as the implicit depth features for subsequent processing. These features encode object layout, relative positional relationships, and cross-view spatial consistency, and are later used by the Spatial Enhancement Module to modulate the 2D vision-language representations.

\noindent\textbf{Vision-Language Backbone.}
The Vision-Language Backbone encodes RGB observations together with the language instruction to produce 2D vision-language representations. Formally, given multi-view RGB observations $\{I_t^i\}_{i=1}^{N}$ and the language instruction $L_t$, the backbone produces
\begin{equation}
Z_t = f_{\mathrm{VLB}}\left(\{I_t^i\}_{i=1}^{N}, L_t\right).
\end{equation}
We instantiate this backbone with a pretrained InternVL3-1B model~\cite{zhu2025internvl3}. It is pretrained under a native multimodal paradigm, in which text corpora and diverse multimodal data are jointly optimized within a single pretraining stage, enabling strong cross-modal alignment between visual observations and task instructions. Its visual encoder adopts InternViT-300M~\cite{chen2024internvl}, while the language branch leverages Qwen2.5-0.5B~\cite{hui2024qwen2}. We retain only the first 14 layers of the language branch, as intermediate layers tend to exhibit stronger cross-modal alignment between visual and linguistic features, making them more suitable for downstream visuomotor control~\cite{shukor2025smolvla}. 
The hidden representations produced by this backbone are taken as the 2D vision-language representations $Z_t$ of the current observation for subsequent spatial enhancement.

\noindent\textbf{Spatial Enhancement Module.}
The Spatial Enhancement Module (SEM) is designed to inject the implicit depth features extracted by the IDEM into the 2D vision-language representations, so that the resulting representations are enriched with spatial information for downstream action prediction.
Given the 2D vision-language representations $Z_t$ and the implicit depth features $D_t$, the SEM first transforms the implicit depth features into a representation aligned with the hidden space of the 2D vision-language representations for feature-wise modulation. 
The transformed depth features are then averaged across tokens to obtain a compact global descriptor $g_t$. Based on this descriptor, a lightweight modulation head $f_{\mathrm{mod}}(\cdot)$ predicts a pair of feature-wise modulation parameters, $\gamma$ and $\beta$:
\begin{equation}
\gamma, \beta = f_{\mathrm{mod}}(g_t),
\end{equation}
where $\gamma$ and $\beta$ are used as channel-wise scaling and shifting factors, respectively.
These parameters are then applied to the 2D vision-language representations in a FiLM-style manner:
\begin{equation}
\hat{Z}_t = \gamma \odot Z_t + \beta,
\end{equation}
where $\odot$ denotes element-wise multiplication. The resulting outputs form the spatially enhanced vision-language representations, which are subsequently passed to the Action Expert for action generation.

\noindent\textbf{Action Expert.}
The Action Expert is instantiated as a conditional denoising policy based on the flow-matching paradigm~\cite{lipman2022flow}, and is implemented with a Diffusion Transformer~\cite{peebles2023scalable} to generate the future action sequence from the spatially enhanced vision-language representations and the current robot state. Specifically, under the joint conditioning of the spatially enhanced vision-language representations $\hat{Z}_t$ and the robot state $s_t$, the module learns a time-dependent vector field that progressively transforms noisy actions toward the target action sequence. Given the ground-truth action sequence $A_t$ and a randomly sampled noise vector $\epsilon$, the noisy action is constructed by linear interpolation:
\begin{equation}
A_t^{\tau} = \tau A_t + (1-\tau)\epsilon,
\end{equation}
where $\tau \in [0,1]$ denotes the interpolation coefficient. The training objective then follows the flow-matching formulation:
\begin{equation}
\mathcal{L}^{\tau}(\theta)
=
\mathbb{E}_{p(A_t \mid \hat{Z}_t, s_t),\, q(A_t^{\tau} \mid A_t)}
\left[
\left\| v_{\theta}(A_t^{\tau}, \hat{Z}_t, s_t) - u(A_t^{\tau} \mid A_t) \right\|^2
\right],
\end{equation}
where $u(A_t^{\tau} \mid A_t)$ denotes the target flow direction toward the ground-truth action sequence. During inference, the policy starts from a random action initialization and iteratively refines it through multiple denoising steps, ultimately producing the final executable robot actions.

\subsection{Progressive Alignment Training}

To better align implicit depth-enhanced spatial representations with downstream action learning, Evo-Depth adopts a progressive alignment training strategy.
This strategy is motivated by two considerations: first, to gradually align newly initialized modules with pretrained components before full end-to-end optimization; second, to encourage more effective integration between implicit depth features and 2D vision-language representations.

Specifically, training is divided into three stages. In Stage 1 (Initial Alignment), we optimize the SEM and Action Expert while freezing the IDEM and the Vision-Language Backbone, establishing an initial alignment between the newly introduced modules and the pretrained representation space. In Stage 2 (Spatial Alignment), we unfreeze the IDEM while keeping the Vision-Language Backbone frozen, enabling implicit depth representations to align more effectively with 2D vision-language features and improving spatial consistency across modules. In Stage 3 (Joint Optimization), all modules are jointly optimized end-to-end, allowing the entire model to achieve full feature coordination and task-specific adaptation.
By progressively establishing alignment across modules, this strategy improves representation compatibility and facilitates deeper fusion between depth-aware and vision-language features, ultimately benefiting the overall action modeling capability. Its effectiveness is further examined through an ablation study in Section~\ref{abl:multistage_train}, and qualitative
visualization in Appendix~\ref{app:idem_vis} further suggests that progressive alignment training encourages more task-focused IDEM spatial attention.

\begin{table*}[t]
  \centering
  \small
  \setlength{\tabcolsep}{4pt}
  \renewcommand{\arraystretch}{0.85}
  \caption{\textbf{Simulation Results.} Comparison across Meta-World,
  VLA-Arena, and LIBERO. ``Params'' denotes model parameters, and ``Robo-
  Pretrain''
  indicates robotic-data pretraining. \textbf{Bold} and \underline{underline} denote the best and
  second-best values within each benchmark; for ``Params'', they denote the
  smallest and second-smallest values. In VLA-Arena, ``Extrap.'' and ``Long''
  denote
  Extrapolation and Long Horizon, respectively.}

  \label{tab:sim_merged_main}
  \resizebox{\textwidth}{!}{%
  \begin{tabular}{l l c c c c c c c}
      \toprule
      \textbf{Benchmark} & \textbf{Models} & \textbf{Params} &
\textbf{Robo-Pretrain} & \multicolumn{5}{c}{\textbf{Success Rate (\%)}} \\
      \midrule
      \textbf{Meta-World} & & & & \textbf{Easy} & \textbf{Medium} &
\textbf{Hard} & \textbf{Very Hard} & \textbf{Avg.} \\
      \midrule
      & TinyVLA~\cite{wen2025tinyvla} & 1.3B & No & 77.6 & 21.5 & 11.4 & 15.8 & 31.6 \\
      & $\pi_0$~\cite{black2410pi0} & 3.5B & Yes & 71.8 & 48.2 & 41.7 & 30.0 & 47.9 \\
      & SmolVLA~\cite{shukor2025smolvla} & 2.3B & No & \underline{87.1} & 51.8 & 70.0 & 64.0 & 68.2
\\
      & RoboTron Mani~\cite{yan2024robotron} & 4B & No & 85.5 & 67.7 & 76.7 & \underline{81.0} &
77.7 \\
      & Evo-1~\cite{lin2025evo1} & \textbf{0.8B} & No & \textbf{89.2} & \underline{76.8} &
\underline{77.2} & 79.2 & \underline{80.6} \\
      & \textbf{Evo-Depth (Ours)} & \underline{0.9B} & No & 83.1 &
\textbf{84.7} & \textbf{87.3} & \textbf{82.4} & \textbf{84.4} \\
      \midrule
      \textbf{VLA-Arena} & & & & \textbf{Safety} & \textbf{Distractor} &
\textbf{Extrap.} & \textbf{Long} & \textbf{Total} \\
      \midrule
      & SmolVLA~\cite{shukor2025smolvla} & \underline{2.3B} & No & 16.5 & 21.0 & 10.4 & 24.7 & 16.4
\\
      & $\pi_0$-FAST~\cite{pertsch2025fast} & 3.5B & Yes & 34.1 & 39.0 & 4.2 & 20.7 & 25.6 \\
      & UniVLA~\cite{bu2025univla} & 7B & Yes & \underline{45.5} & 41.3 & \underline{31.1} &
22.0 & 38.7 \\
      & OpenVLA~\cite{kim2024openvla} & 7B & Yes & 41.9 & 43.0 & \textbf{37.8} & \underline{26.7}
& 39.6 \\
      & OpenVLA-OFT~\cite{kim2025fine} & 7B & Yes & 43.2 & \underline{52.3} & 30.4 &
\underline{26.7} & \underline{39.9} \\
      & \textbf{Evo-Depth (Ours)} & \textbf{0.9B} & No & \textbf{47.3} &
\textbf{52.7} & 25.8 & \textbf{46.5} & \textbf{41.1} \\
      \midrule
      \textbf{LIBERO} & & & & \textbf{Spatial} & \textbf{Object} &
\textbf{Goal} & \textbf{Long} & \textbf{Avg.} \\
      \midrule
      & TraceVLA~\cite{zheng2024tracevla} & 7B & Yes & 84.6 & 85.2 & 75.1 & 54.1 & 74.8 \\
      & SpatialVLA~\cite{qu2025spatialvla} & 4B & Yes & 88.2 & 89.9 & 78.6 & 55.5 & 78.1 \\
      & 4D-VLA~\cite{zhang20254d} & 4B & Yes & 88.9 & 95.2 & 90.9 & 79.1 & 88.6 \\
      & SmolVLA~\cite{shukor2025smolvla} & 2.3B & No & 93.0 & 94.0 & 91.0 & 77.0 & 88.8 \\
      & GR00T-N1~\cite{bjorck2025gr00t} & 2B & Yes & 94.4 & 97.6 & 93.0 & 90.6 & 93.9 \\
      & DM0~\cite{yu2026dm0} & 2.4B & Yes & \textbf{98.2} & 98.8 & \textbf{96.6} & 82.6 &
94.1 \\
      & $\pi_0$~\cite{black2410pi0} & 3.5B & Yes & 96.8 & 98.8 & 95.8 & 85.2 & 94.2 \\
      & $\pi_{0.5}$~\cite{intelligence2025pi_} & 3.5B & Yes & \underline{98.0} & 97.8 & 95.6 & 85.8 &
94.3 \\
      & VLA-0~\cite{goyal2025vla} & 3B & No & 97.0 & 97.8 & 96.2 & 87.6 & 94.7 \\
      & BitVLA~\cite{wang2025bitvla} & 3B & Yes & 97.4 & \textbf{99.6} & 94.4 & 87.7 & 94.8 \\
      & Evo-1~\cite{lin2025evo1} & \textbf{0.8B} & No & 92.7 & 97.7 & \underline{96.3} &
\textbf{92.3} & 94.8 \\
      & DepthVLA~\cite{yuan2025depthvla} & 4.1B & No & 96.4 & 98.0 & 95.8 & 89.2 & 94.9 \\
      & NORA-1.5~\cite{hung2025nora} & 3B & Yes & \underline{98.0} & 96.0 & 95.4 & 90.5 & 95.0
\\
      & UniVLA~\cite{bu2025univla} & 7B & Yes & 96.5 & 96.8 & 95.6 & \underline{92.0} &
\underline{95.2} \\
      & \textbf{Evo-Depth (Ours)} & \underline{0.9B} & No & 95.6 &
\underline{99.2} & 95.6 & 91.3 & \textbf{95.4} \\
      \bottomrule
  \end{tabular}%
  }
\end{table*}

\begin{table}[t]
    \centering
    \small
    \caption{\textbf{LIBERO-Plus Results.} Results in LIBERO-Plus benchmark. ``Bkgd.'' denotes Background. \textbf{Bold} indicates the best result in each column, and \underline{underlined} indicates the second-best result.}
    \label{tab:libero_plus}
    \setlength{\tabcolsep}{3.5pt}
    \renewcommand{\arraystretch}{0.8}
    \resizebox{\columnwidth}{!}{%
    \begin{tabular}{>{\raggedright\arraybackslash}p{2.9cm}*{8}{>{\centering\arraybackslash}p{1.05cm}}}
        \toprule
        Model & Camera & Robot & Language & Light & Bkgd. & Noise & Layout & Avg. \\
        \midrule
        WorldVLA~\cite{cen2025worldvla} & 0.1 & 27.9 & 41.6 & 43.7 & 17.1 & 10.9 & 38.0 & 25.0 \\
        NORA~\cite{hung2025nora} & 2.2 & 37.0 & 65.1 & 45.7 & 58.6 & 12.8 & 62.1 & 39.0 \\
        UniVLA~\cite{bu2025univla} & 1.8 & \underline{46.2} & \underline{69.6} & 69.0 & \underline{81.0} & 21.2 & 31.9 & 43.9 \\
        $\pi_0$-Fast~\cite{intelligence2025pi_} & \underline{13.8} & 6.0 & 58.8 & \underline{85.0} & \textbf{81.4} & \textbf{79.0} & \underline{68.9} & \underline{53.6} \\
        \textbf{Evo-Depth (Ours)} & \textbf{47.2} & \textbf{49.2} & \textbf{78.9} & \textbf{88.1} & 76.4 & \underline{77.6} & \textbf{69.6} & \textbf{69.6} \\
        \toprule
    \end{tabular}
    }
    \renewcommand{\arraystretch}{1}
\end{table}
\section{Simulation Experiments}

\noindent \textbf{Setup.}
We evaluate Evo-Depth on four representative simulation benchmarks, including
Meta-World~\cite{yu2020meta}, VLA-Arena~\cite{zhang2025vla},
LIBERO~\cite{liu2023libero}, and LIBERO-Plus~\cite{fei2025libero}, to assess
manipulation performance under varying task difficulty, long-horizon
execution, and distribution shifts. Detailed benchmark settings and evaluation
protocols are provided in Appendix~\ref{app:sim_setup}.
   
\noindent \textbf{Results.}
As shown in Table~\ref{tab:sim_merged_main}, Evo-Depth achieves state-of-the-art overall performance on Meta-World, obtaining the highest average success rate of 84.4 and the best results on the \textit{medium}, \textit{hard}, and \textit{very hard} splits. On VLA-Arena, Evo-Depth delivers strong overall performance, with particularly good results in the \textit{distractor} and \textit{long horizon} settings. On LIBERO, Evo-Depth also performs strongly, achieving a high average success rate of 95.4 with balanced performance across different task suites. As shown in Table~\ref{tab:libero_plus}, Evo-Depth further demonstrates strong robustness on LIBERO-Plus, achieving impressive overall average and performing well across diverse perturbation dimensions. Overall, these results demonstrate that the implicit depth perception strategy effectively improves VLA performance across diverse simulation benchmarks, especially in challenging scenarios that require robust spatial perception, long-horizon execution, and generalisation under distribution shifts.

\section{Real-World Experiments}

\begin{figure}[t]
    \centering
    \includegraphics[width=\linewidth]{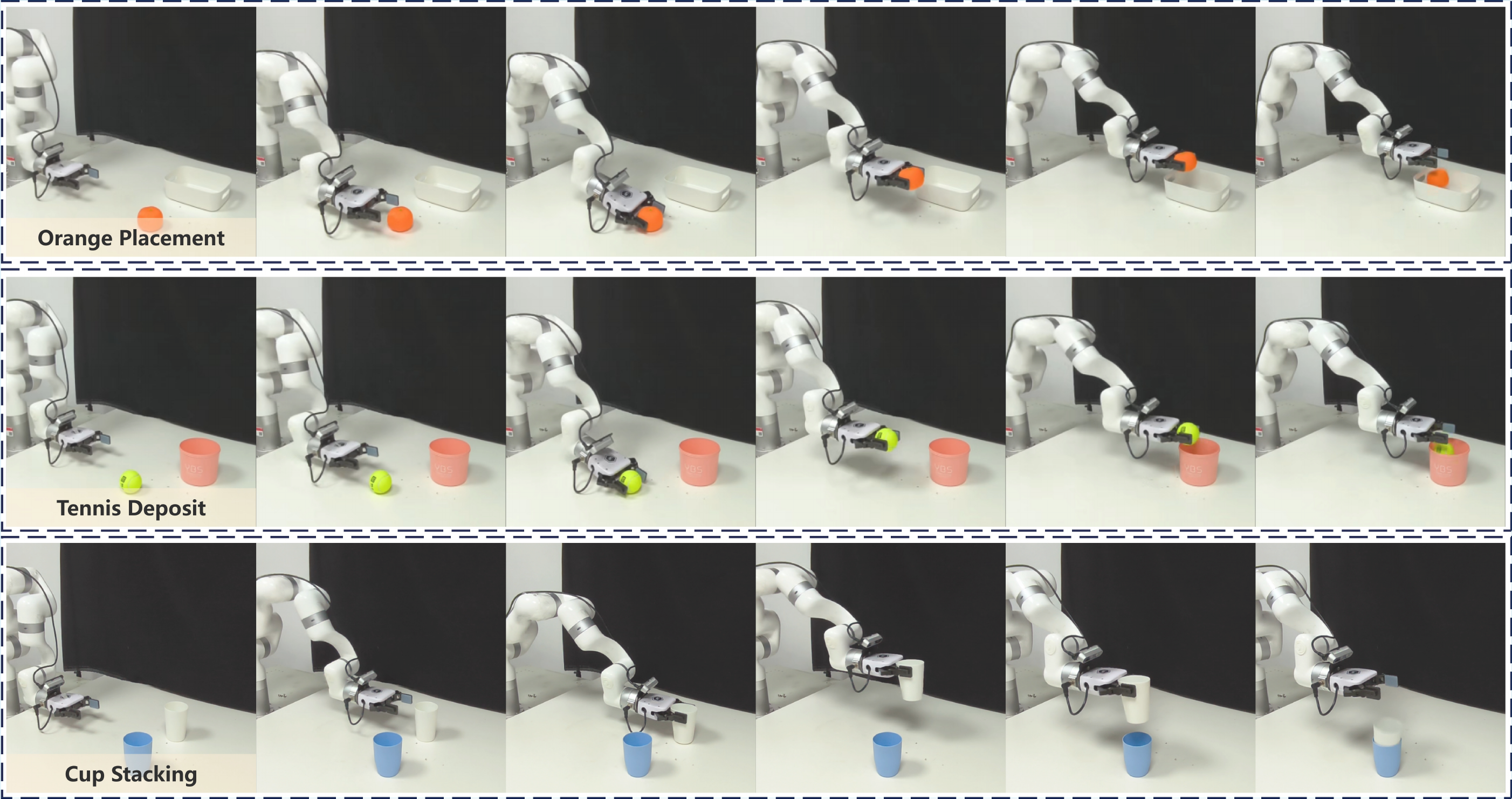}
    \caption{\textbf{Real-World Setup.} Real-world experiment process for three tasks, \textit{Orange Placement}, \textit{Tennis Deposit}, and \textit{Cup Stacking}, with increasing manipulation difficulty.}
    \label{fig:real_world_setup}
\end{figure}

\noindent\textbf{Setup.}
We evaluate Evo-Depth on a physical robot across three real-world manipulation
tasks of increasing difficulty, as illustrated in
Fig.~\ref{fig:real_world_setup}: \textit{Orange Placement}, \textit{Tennis
Deposit}, and \textit{Cup Stacking}. These tasks impose progressively stronger
requirements on spatial perception and fine-grained manipulation. Notably,
the third-view camera is partially occluded during critical manipulation
stages, making the setup particularly challenging for accurate grasping and
placement. For each task, we conduct 20 trials, and the success rate is
computed as the ratio of successful trials to the total number of trials.
Detailed task definitions and real-world evaluation protocols are provided in
Appendix~\ref{app:real_setup}.

\noindent\textbf{Results.}
Evo-Depth achieves the strongest overall performance across all three real-world tasks. As shown in Table~\ref{tab:inference_efficiency}, the performance gap among the three methods becomes more evident on tasks with higher spatial complexity. In particular, tasks involving target motion and precise object-object interaction are more sensitive to the model's spatial perception quality. This trend suggests that real-world manipulation performance is strongly related to how well a model can capture object layout and relative positions from visual observations. The results further indicate that introducing implicit depth cues into VLA learning can provide more informative spatial representations, which is beneficial for more accurate action generation in challenging real-world settings. We also conduct a
detailed failure case analysis for the three real-world tasks, which is provided in the Appendix~\ref{app:failure_cases}.

\noindent\textbf{Inference Efficiency Analysis.}
We further compare baseline VLA models in terms of parameter size, GPU memory usage, inference frequency, and average success rate in real-robot experiments. As shown in Table~\ref{tab:inference_efficiency}, Evo-Depth achieves the best overall trade-off between efficiency and performance. Despite using only 0.9B parameters and 3.2 GB GPU memory, Evo-Depth runs at 12.3 Hz and attains an average success rate of 90.0\%, significantly outperforming OpenVLA-OFT and $\pi_0$ in real-world manipulation tasks.

\begin{table*}[t]
\centering
\caption{\textbf{Real-Robot Comparison.} Inference efficiency and task-wise
success rate comparison of different VLA models in real-robot experiments.
\textbf{Bold} and \underline{underline} denote the best and second-best values, respectively,
except for ``Params'' and ``GPU Mem.'', where they denote the smallest and
second-smallest values. All task-wise success rates and average success are reported in percent.}
\label{tab:inference_efficiency}
\renewcommand{\arraystretch}{}
\setlength{\tabcolsep}{4pt}
\resizebox{\textwidth}{!}{
  \begin{tabular}{lccccccc}
  \toprule
  Model & Params (B) & GPU Mem. (GB) & Infer. Freq. (Hz) & Orange Placement &
  Tennis Deposit &
  Cup Stacking & Avg. Success (\%) \\
  \midrule
  OpenVLA-OFT~\cite{kim2025fine}      & 7.0 & \underline{15.1} & 7.8  & 50.0 &
  40.0 & 25.0 & 38.3
  \\
  $\pi_0$~\cite{black2410pi0}          & \underline{3.5} & 17.9 &
  \underline{10.2} &
  \underline{75.0} & \underline{70.0} & \underline{55.0} & \underline{66.7} \\
  \textbf{Evo-Depth (Ours)} & \textbf{0.9} & \textbf{3.2}  & \textbf{12.3} &
  \textbf{95.0} & \textbf{90.0} & \textbf{85.0} & \textbf{90.0} \\
  \bottomrule
  \end{tabular}

}
\end{table*}

\section{Generalization Experiments}

\begin{figure}[t]
    \centering
    \includegraphics[width=\linewidth]{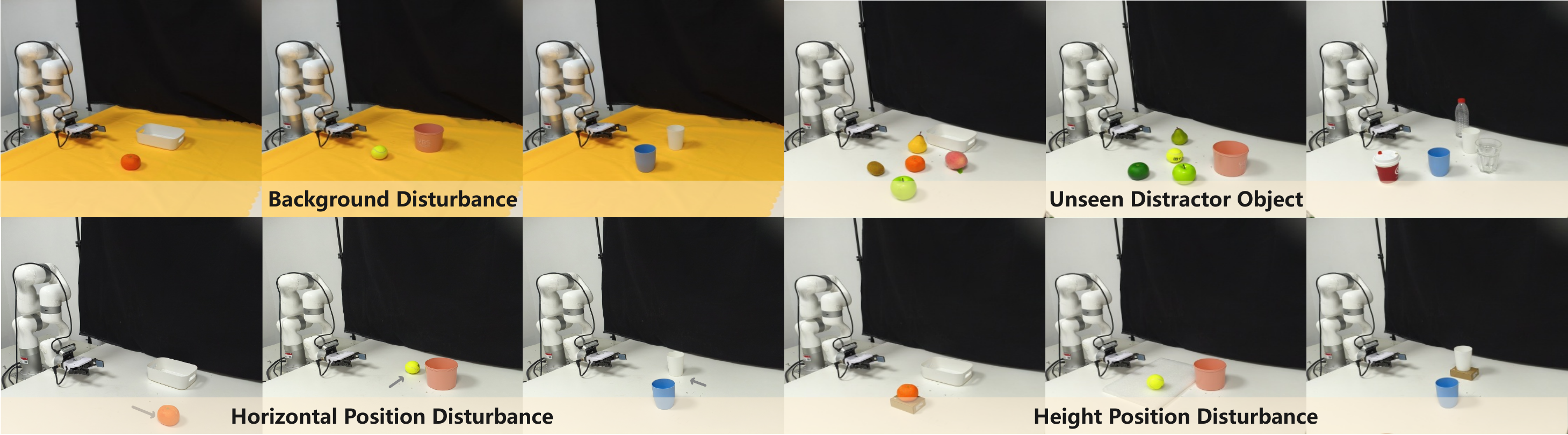}
    \caption{\textbf{Generalization Setup.} Generalization experiments under four disturbance settings: background disturbance, unseen distractors, horizontal and height position disturbance.}
    \label{fig:generalization_setup}
\end{figure}

\noindent\textbf{Setup.}
To evaluate the real-world generalization ability of Evo-Depth, we further
test it under four types of distribution shifts and environmental disturbances
using the same three tasks as in the real-world experiments: \textit{Orange
Placement}, \textit{Tennis Deposit}, and \textit{Cup Stacking}. As illustrated
in Fig.~\ref{fig:generalization_setup}, these settings cover appearance
changes, distractor interference, and spatial perturbations in both horizontal
and vertical dimensions. For each task under each disturbance setting, we
conduct 20 trials, and the success rate is computed as the ratio of successful
trials to the total number of trials. Detailed disturbance settings are
provided in Appendix~\ref{app:gen_setup}.
  
\noindent\textbf{Results.}
As shown in Fig.~\ref{fig:generation_results}, Evo-Depth consistently achieves higher success rates than the baseline under all four disturbance settings across the three real-world tasks.
These results demonstrate that Evo-Depth generalizes more effectively to unseen visual and spatial variations. In particular, its advantage becomes more significant in challenging settings that require accurate spatial reasoning, such as position perturbations and precise object-object interaction. This suggests that implicit depth-aware representations enhance spatial perception and improve the robustness of VLA policies under real-world distribution shifts.

\section{Ablation Studies}

\subsection{Implicit Depth Perception Analysis}

\textbf{Ablation Setup.}
To evaluate the contribution of implicit depth perception, we remove the
implicit depth perception module from Evo-Depth while keeping the remaining
architecture and training pipeline unchanged. We conduct this ablation on
LIBERO-Plus, and detailed evaluation settings are provided in
Appendix~\ref{app:ablation_setup}.

\textbf{Results Analysis.}
As shown in Fig.~\ref{fig:idem_ablation}, Evo-Depth consistently outperforms the variant without the implicit depth perception module across all four evaluation dimensions. The gains are particularly clear on \textit{camera}, \textit{robot}, and \textit{layout}, indicating that implicit depth cues are especially helpful for handling viewpoint changes, embodiment variation, and object arrangement. Overall, these results verify the effectiveness of implicit depth perception in improving spatial perception and manipulation robustness.

\subsection{Multi-Stage Fusion Training}
\label{abl:multistage_train}
\noindent\textbf{Ablation Setup.}
To evaluate the proposed Progressive Alignment Training strategy, we compare
one-stage, two-stage, and three-stage training variants on the LIBERO Long
benchmark. Detailed stage definitions are provided in
Appendix~\ref{app:ablation_setup}.
  
\begin{figure}[t]                                                             
  \centering
  \captionsetup[subfigure]{font=footnotesize, skip=2pt}
  \begin{subfigure}[t]{0.34\linewidth}                                      
      \centering                                                         
      \includegraphics[width=\linewidth]{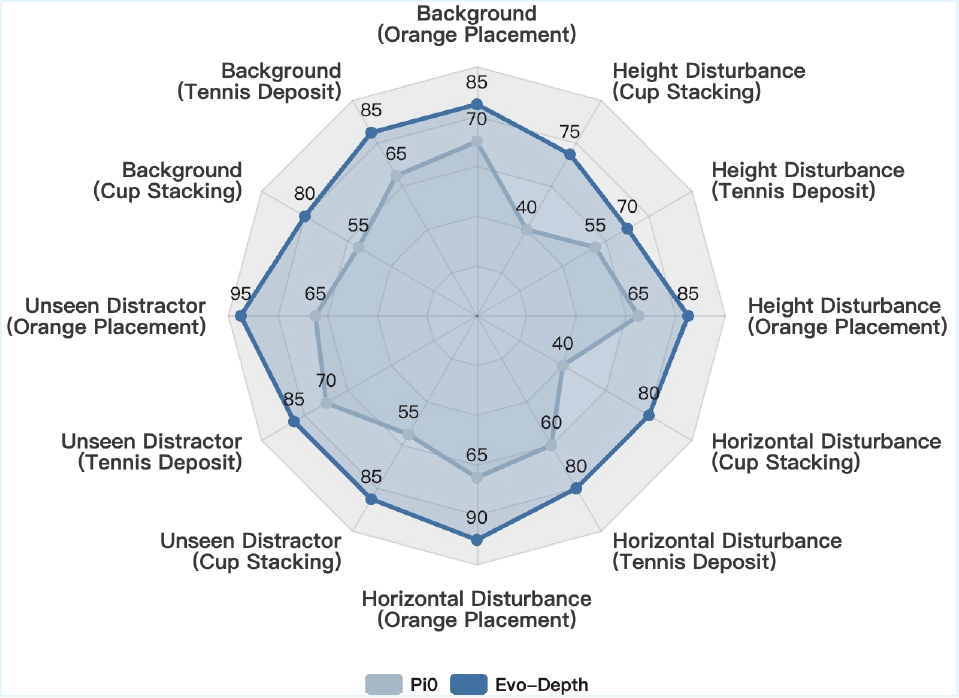}
      \caption{}
      \label{fig:generation_results}
  \end{subfigure}
  \hfill
  \begin{subfigure}[t]{0.33\linewidth}
      \centering
      \includegraphics[width=\linewidth]{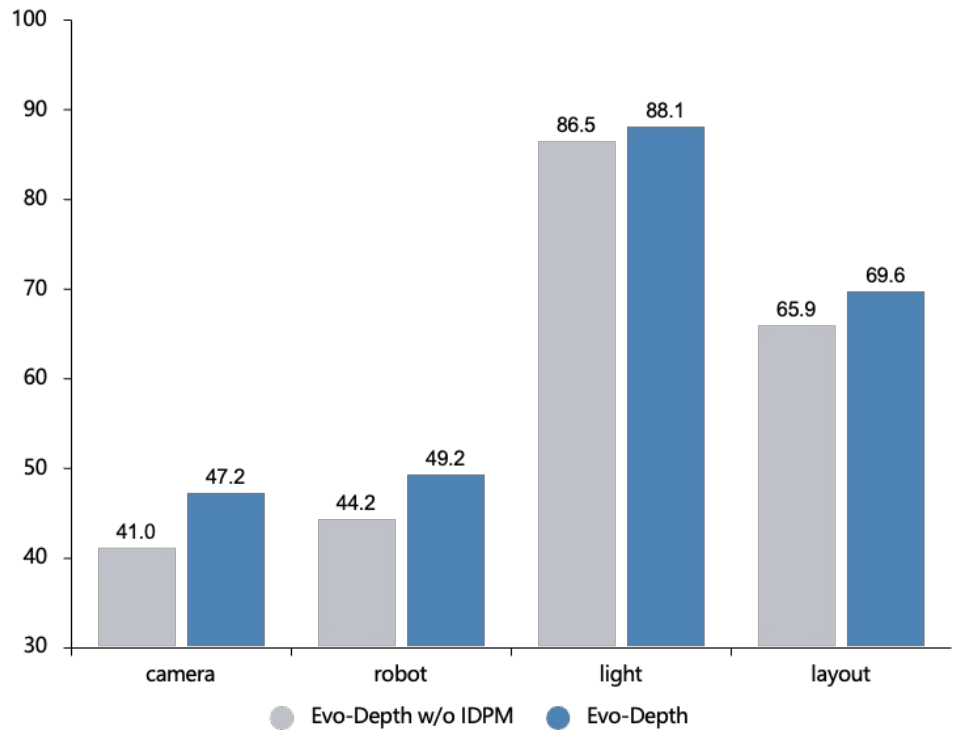}
      \caption{}
      \label{fig:idem_ablation}
  \end{subfigure}
  \hfill
  \begin{subfigure}[t]{0.15\linewidth}
      \centering
      \includegraphics[width=\linewidth]{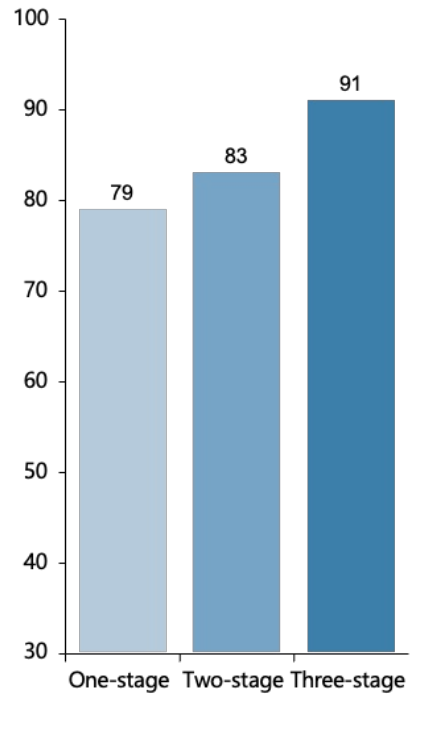}
      \caption{}
      \label{fig:alignment_ablation}
  \end{subfigure}
  \hfill
  \begin{subfigure}[t]{0.15\linewidth}
      \centering
      \includegraphics[width=\linewidth]{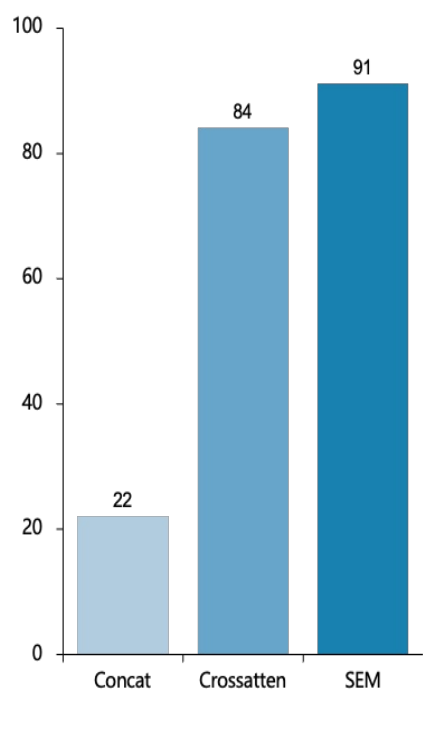}
      \caption{}
      \label{fig:fusion_ablation}
  \end{subfigure}
  \caption{\textbf{Generalization and Ablation Studies on Evo-Depth.} (a) Generalization performance of Pi0 and Evo-Depth across multiple perturbations. (b) Effect of the implicit depth perception module on LIBERO-Plus under four evaluation dimensions. (c) Effect of different alignment training stages on LIBERO Long benchmark. (d) Effect of different fusion strategies on LIBERO Long benchmark.}
  \label{fig:multi_stage_ablation}
\end{figure}

\textbf{Results Analysis.}
The results consistently verify the effectiveness of the proposed Progressive Alignment Training strategy. As shown in Fig.~\ref{fig:alignment_ablation}, the three-stage variant achieves the best performance, outperforming both the one-stage and two-stage variants. This result suggests that implicit depth cues are more effectively incorporated into the vision-language-action framework when cross-module alignment is introduced progressively rather than through abrupt joint optimization. In particular, while the two-stage variant already improves over one-stage training, the additional intermediate alignment stage in the three-stage variant further strengthens multimodal integration and leads to the best overall performance.

\subsection{Fusion Strategy Analysis}

\noindent\textbf{Ablation Setup.}
To investigate the impact of different fusion strategies, we compare
concatenation, cross-attention, and SEM-based modulation for integrating
implicit depth features into the 2D vision-language representations on the LIBERO Long benchmark. Detailed
variant definitions are provided in Appendix~\ref{app:ablation_setup}.
  
\textbf{Results Analysis.}
As shown in Fig.~\ref{fig:fusion_ablation}, different fusion strategies lead to markedly different performance. The \textit{Concat} variant performs substantially worse than the other two variants, suggesting that treating implicit depth features and 2D vision-language representations in a symmetric manner is not effective for downstream action prediction. In contrast, both \textit{Crossatten} and \textit{SEM} achieve much stronger results, indicating that implicit depth cues are more useful when they are incorporated as complementary spatial signals conditioned on the 2D vision-language representations. Among them, \textit{SEM} achieves the best performance, suggesting that FiLM-style feature-wise modulation provides a more effective way to inject spatial information while preserving the dominant role of the original 2D vision-language representations.

\section{Conclusion}
We introduced Evo-Depth, a lightweight depth-enhanced VLA framework that improves spatially grounded robotic manipulation while preserving deployment efficiency without relying on additional sensing hardware. Evo-Depth integrates an Implicit Depth Encoding Module, a Spatial Enhancement Module, and Progressive Alignment Training to enhance depth-aware spatial understanding for downstream action learning. Experiments in simulation and real-world settings demonstrate its effectiveness, robustness, and deployment efficiency. These results suggest that lightweight implicit depth enhancement is an effective direction for practical VLA systems. 

\newpage
\bibliographystyle{plainnat}
\bibliography{references}

\newpage
\appendix
\section{Simulation Experiment Details}

\subsection{Simulation Benchmark Setup}
\label{app:sim_setup}
\begin{figure}[t]
    \centering
    \includegraphics[width=\linewidth]{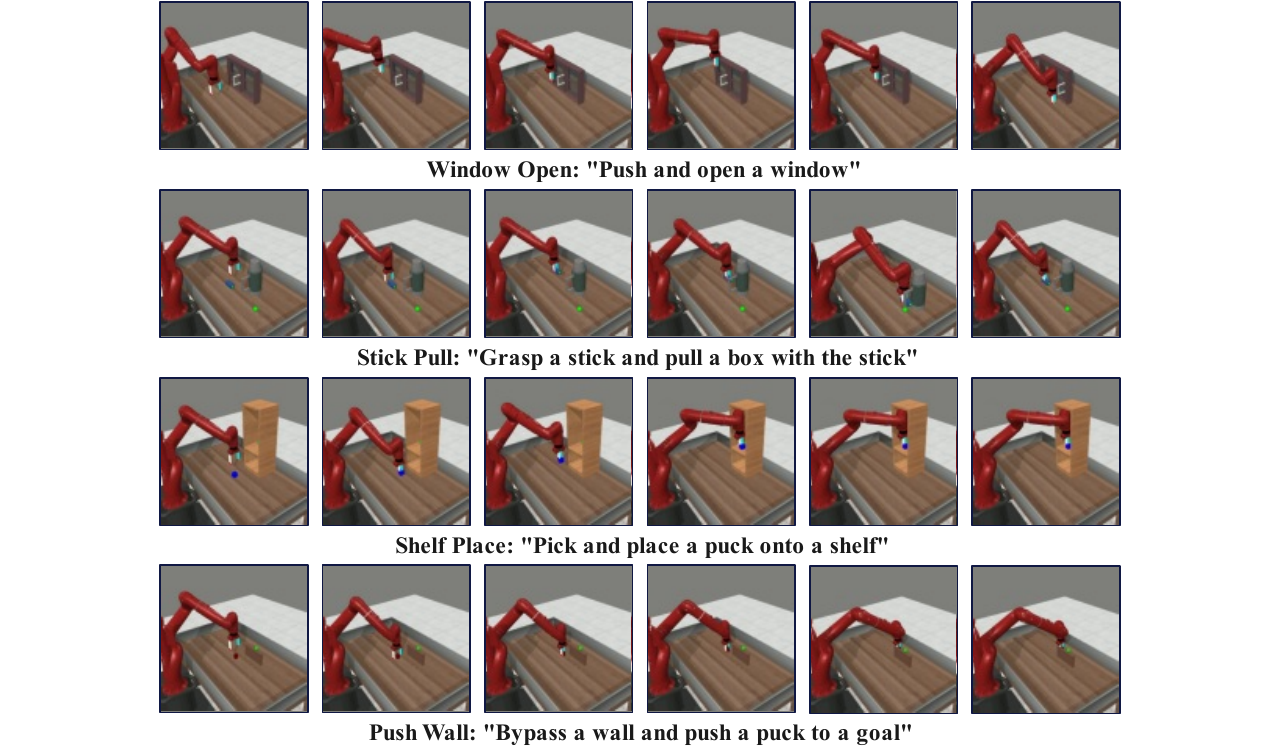}
    \caption{\textbf{Representative task trajectories from Meta-World.} We show example multi-step manipulation sequences from several Meta-World tasks to illustrate the benchmark's diversity in object interaction and spatial control.}
    \label{fig:appendix_metaworld}
\end{figure}

\begin{figure}[t]
    \centering
    \includegraphics[width=\linewidth]{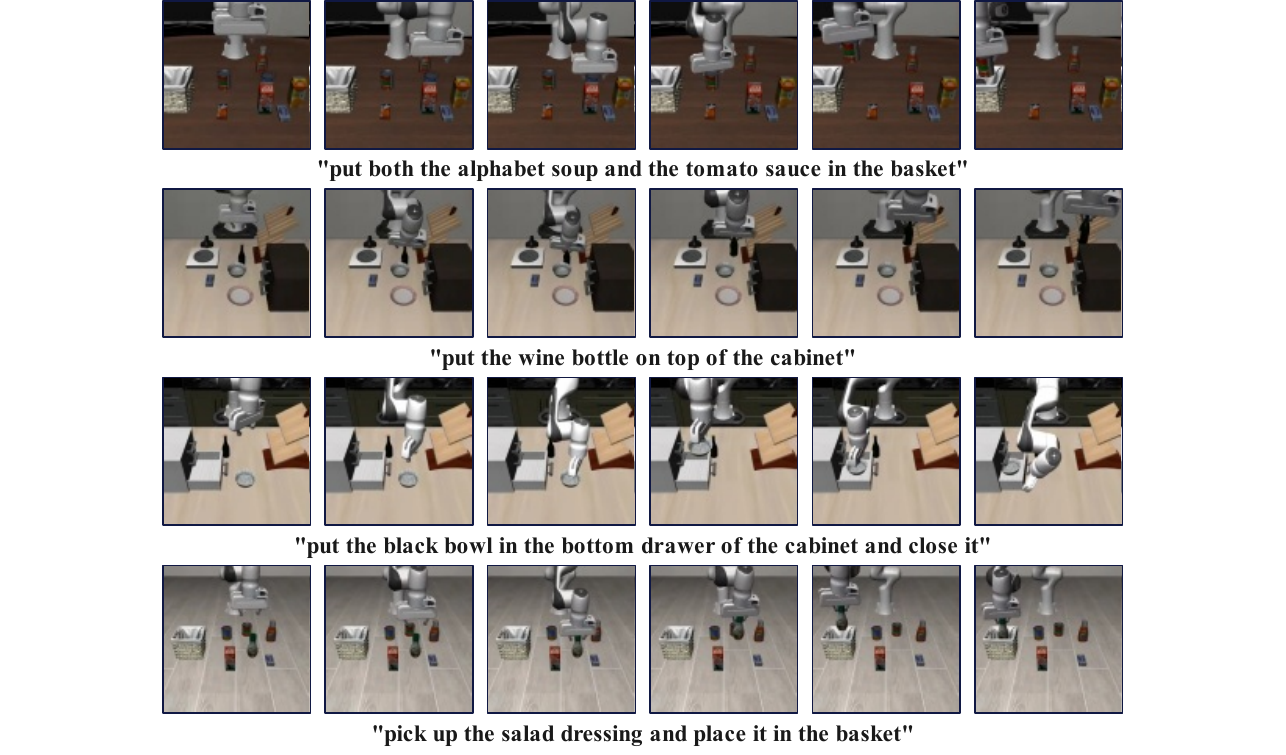}
    \caption{\textbf{Representative task trajectories from LIBERO.} We show example task sequences from LIBERO to illustrate the benchmark's language-conditioned manipulation settings and varied object arrangements.}
    \label{fig:appendix_libero}
\end{figure}

We evaluate Evo-Depth on four complementary simulation benchmarks to assess
its effectiveness from different perspectives, including base manipulation
competence, VLA-oriented generalization, cross-task transfer, and robustness
under perturbations.
For all benchmarks, we report success rates and compare Evo-Depth with
representative VLA baselines under aligned evaluation settings. All baseline results are derived from the original paper or official implementation. Representative task trajectories from Meta-World and LIBERO are shown in
Figs.~\ref{fig:appendix_metaworld} and~\ref{fig:appendix_libero},
respectively, to illustrate the diversity of manipulation scenes considered in
these benchmarks.

\noindent\textbf{Meta-World.}
Meta-World is a widely used multi-task robotic manipulation benchmark that
contains 50 distinct manipulation environments. Its
task set is intentionally diverse yet structured, covering a broad range of
fundamental manipulation skills with shared underlying object-interaction
patterns. We include Meta-World because it provides a broad evaluation of
whether Evo-Depth improves core manipulation ability beyond narrow single-task
settings. In our evaluation, we group the benchmark into four difficulty splits
(\textit{easy}, \textit{medium}, \textit{hard}, and \textit{very hard}) to
analyze how performance changes as tasks demand more accurate localization,
spatial coordination, and fine-grained object placement. This benchmark is
particularly relevant to our study because it tests whether lightweight
implicit depth cues remain beneficial as manipulation difficulty increases.

\noindent\textbf{VLA-Arena.}
VLA-Arena is a benchmark designed specifically for evaluating
Vision-Language-Action models. Its original design quantifies VLA difficulty
along three orthogonal axes, namely task structure, language command, and
visual observation, and organizes 170 tasks into four task-structure
dimensions: \textit{safety}, \textit{distractor}, \textit{extrapolation}, and
\textit{long horizon}.
In particular, the \textit{distractor} and \textit{long horizon} dimensions
are closely related to the ability to maintain stable spatial grounding over
cluttered observations and multi-stage action sequences, while
\textit{extrapolation} probes generalization beyond seen conditions. This
benchmark is particularly relevant to our study because it directly tests
whether the proposed implicit-depth enhancement improves robustness and spatial
reasoning in challenging VLA settings without sacrificing the lightweight
nature of the policy.

\noindent\textbf{LIBERO.}
LIBERO is used in our evaluation through four categories,
\textit{spatial}, \textit{object}, \textit{goal}, and \textit{long}, which
respectively cover manipulation settings with different emphases on spatial
reasoning, object understanding, goal-conditioned execution, and long-horizon
task composition. We include LIBERO because it provides a balanced evaluation
across multiple types of language-conditioned manipulation settings, rather
than emphasizing only a single source of difficulty. This benchmark is
particularly relevant to our study because it allows us to examine model
behavior under different task demands using a unified evaluation setup. In
particular, the \textit{spatial} and \textit{long} categories are useful for
analyzing settings that place relatively higher demands on spatial grounding
and sustained action coordination over time.

\noindent\textbf{LIBERO-Plus.}
LIBERO-Plus extends LIBERO with a dedicated robustness benchmark for VLA
models. It systematically introduces perturbations along seven dimensions,
including \textit{objects layout}, \textit{camera viewpoints}, \textit{robot
initial states}, \textit{language instructions}, \textit{light conditions},
\textit{background textures}, and \textit{sensor noise}, in order to expose
failure modes that may be hidden by standard benchmark scores. We include
LIBERO-Plus because one of the main motivations of Evo-Depth is to strengthen
spatial perception from RGB alone in a lightweight manner, and such an
improvement should be reflected not only in in-distribution success rates but
also in robustness to visual and spatial perturbations. In our evaluation, we
report results on \textit{camera}, \textit{robot}, \textit{language},
\textit{light}, \textit{background}, \textit{noise}, and \textit{layout}.
This benchmark is particularly relevant to our study because perturbations in
viewpoint, embodiment, and scene arrangement directly test whether the
implicit-depth features extracted by IDEM provide more stable spatial grounding
than a purely 2D visual representation.

\section{Real-World Experiment Details}
\subsection{Real-World Experiment Setup}
\label{app:real_setup}

\noindent\textbf{Task Definitions.}
We evaluate Evo-Depth on three real-world manipulation tasks with increasing
difficulty. \textit{Orange Placement} requires grasping an orange and placing
it into a white box. \textit{Tennis Deposit} involves picking up a tennis ball
and depositing it into a small orange bucket, which is more challenging
because the ball is easily disturbed and may roll away during contact.
\textit{Cup Stacking} requires stacking a white cup onto a blue cup and
demands the most precise spatial understanding and fine-grained control due to
the small geometric tolerance during insertion. Together, these tasks impose
progressively stronger requirements on spatial perception and manipulation.

\begin{figure}[t]
    \centering
    \includegraphics[width=\linewidth]{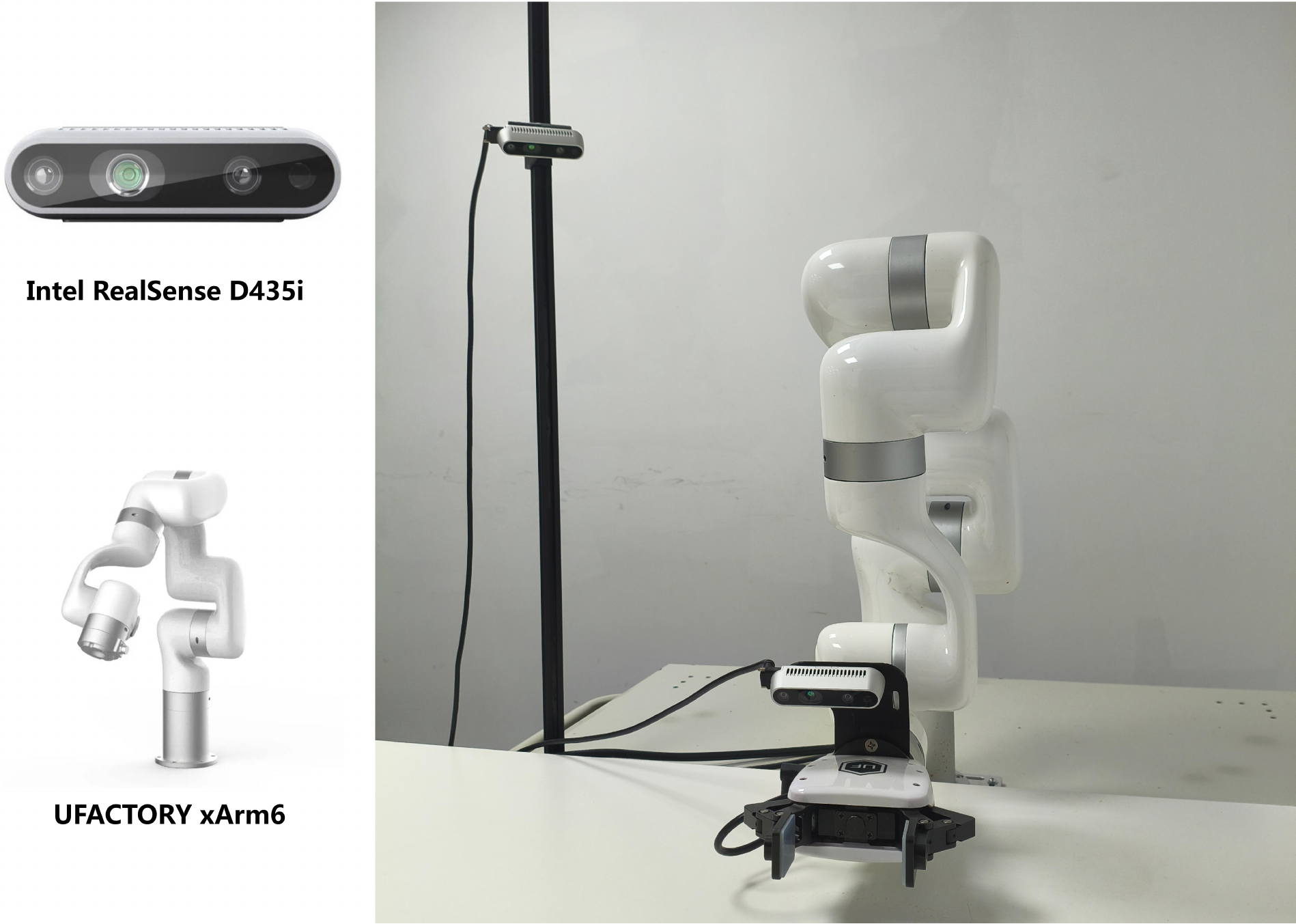}
    \caption{\textbf{Real-world hardware setup.} The real-world platform uses a fixed-base UFACTORY xArm6 robot with two Intel RealSense D435i RGB cameras, including one external camera and one wrist-mounted camera.}
    \label{fig:real_hardware_setup}
\end{figure}

\noindent\textbf{Hardware Setup.}
The real-world experiments are conducted on a fixed-base UFACTORY xArm6
platform. For visual observation, we use two Intel RealSense D435i RGB
cameras, including one external camera mounted on a stand and one
wrist-mounted camera attached near the end effector, as shown in
Fig.~\ref{fig:real_hardware_setup}. This setup provides complementary views
of the workspace and the manipulation area.

\noindent\textbf{Camera Configuration.}
The real-world setup uses two RGB views for observation, including one
external scene camera and one wrist-mounted camera. During critical
manipulation stages, the external scene camera is partially occluded,
resulting in incomplete visual observations when accurate grasping and
placement are most needed. This setting increases the difficulty of real-world
manipulation and provides a challenging testbed for evaluating the spatial
perception and reasoning capability of VLA models.

\noindent\textbf{Evaluation Protocol.}
For each task, we execute the policy under the same observation and control
setting and report task success rates. All compared methods are evaluated
under aligned real-world conditions to ensure fair comparison. Success is
determined by whether the robot completes the corresponding manipulation goal
for each task.

\noindent\textbf{Efficiency Measurement.}
We additionally compare parameter size, GPU memory usage, and inference
frequency in the real-robot setting. GPU memory usage and inference frequency
are measured on an NVIDIA RTX 4090D platform. These metrics are used together
with task success rate to evaluate the deployment efficiency of different VLA
models in real-world manipulation.

\subsection{Failure Case Analysis}
\label{app:failure_cases}
We further analyze the typical failure cases in the three real-world tasks. In \textit{Orange Placement}, failures mainly occur during grasping under partial camera occlusion, where the gripper may collide with the orange and trigger collision protection. In \textit{Tennis Deposit}, failures are primarily caused by the tennis ball being easily disturbed and rolling away when the gripper fails to enclose it securely. In \textit{Cup Stacking}, failures mainly result from the small tolerance between the two cups, where slight misalignment can lead to collision and tipping during insertion. These cases suggest that, although Evo-Depth improves spatial perception and manipulation capability, highly precise contact-rich interactions under occlusion, target motion, and tight geometric constraints remain challenging in real-world settings.

\subsection{Visualization of IDEM Spatial Attention}
\label{app:idem_vis}
\begin{figure*}[t]
    \centering
    \includegraphics[width=\linewidth]{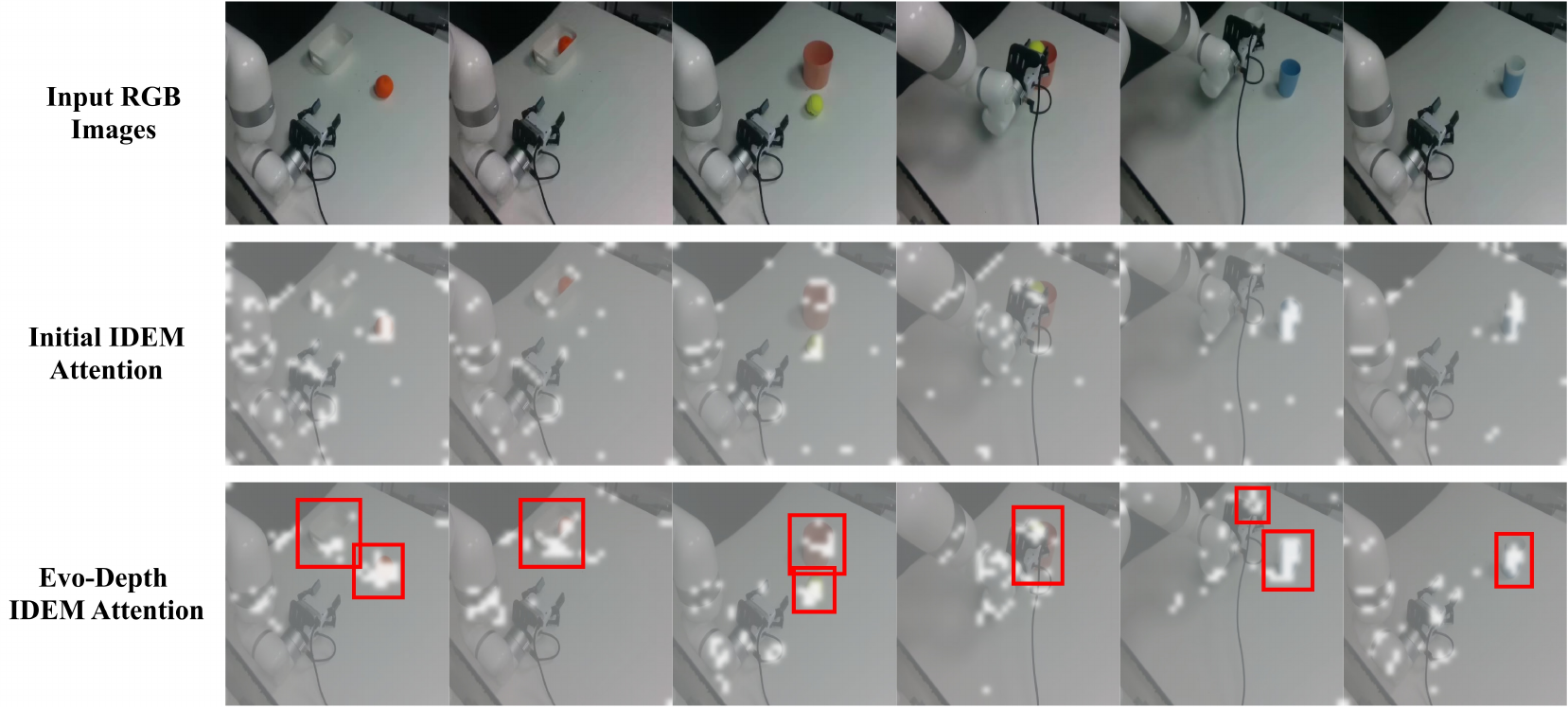}
    \caption{\textbf{Visualization of IDEM spatial attention.} The first row shows input RGB observations from representative stages of real-world manipulation. The second row shows attention maps from the initial IDEM initialized from the depth encoder. The third row shows attention maps from the IDEM in Evo-Depth after progressive alignment training. Compared with the initial IDEM, the trained IDEM exhibits more focused responses on the gripper-object interaction region and the target placement area, suggesting improved alignment between implicit depth features and task-relevant spatial cues.}
    \label{fig:idem_attention_vis}
\end{figure*}

To provide qualitative evidence on how the proposed training strategy affects spatial perception, we visualize the spatial attention patterns of IDEM on representative real-world manipulation frames. As shown in Fig.~\ref{fig:idem_attention_vis}, the initial IDEM already captures coarse object-level structure, while the IDEM in Evo-Depth exhibits more concentrated responses around manipulation-relevant regions after progressive alignment training. In particular, the attention becomes more focused on the gripper-object interaction area and the target placement region. These visualizations provide qualitative support that Evo-Depth encourages implicit depth features to better align with downstream manipulation needs.

\section{Generalization Experiment Details}
\subsection{Generalization Experiment Setup}
\label{app:gen_setup}

We evaluate real-world generalization under four disturbance settings using
the same three tasks as in the main real-world experiments: \textit{Orange
Placement}, \textit{Tennis Deposit}, and \textit{Cup Stacking}. These
disturbances are designed to cover appearance changes, clutter interference,
and spatial perturbations in both horizontal and vertical directions. For each
disturbance setting, the task objective and evaluation criterion remain the
same as in the standard real-world setup, while only the corresponding
environmental factor is changed. Fig.~\ref{fig:generalization_setup}
illustrates representative examples of the four disturbance settings across
the three tasks.

\noindent\textbf{Background Disturbance.}
To introduce an appearance change without altering the task itself, we place a
yellow tablecloth on the work surface while keeping the task objects and goal
configuration unchanged. This setting changes the visual appearance of the
scene and tests whether the policy relies too strongly on the original scene
color and background cues.

\noindent\textbf{Unseen Distractor Object.}
We add several task-irrelevant objects into the workspace, making the scene
more cluttered and increasing visual interference around the target objects.
This setting tests whether the policy can maintain stable attention to the
task-relevant objects and preserve correct action generation when additional
objects are present in the scene.

\noindent\textbf{Horizontal Position Disturbance.}
We perturb the horizontal position of the target object or related task object
within the workspace, so that the relative left-right or front-back spatial
relationship differs from the standard setup. This setting tests whether the
policy can adapt to horizontal spatial changes and still produce accurate grasping
and placement behaviors when the object arrangement is shifted.

\noindent\textbf{Height Position Disturbance.}
We vary the height of the target object or related task object, resulting in a
different vertical spatial relationship from the standard setup. This setting
is intended to test whether the policy can handle changes in object height
during grasping, alignment, and placement, which require more precise spatial
judgment in the vertical direction.

\section{Ablation Study Details}
\subsection{Ablation Setup}
\label{app:ablation_setup}

\noindent\textbf{Implicit Depth Perception Ablation.}
To evaluate the contribution of implicit depth perception, we remove the
implicit depth perception module from Evo-Depth while keeping the remaining
architecture and training pipeline unchanged. We conduct this ablation on
LIBERO-Plus and report results on four representative dimensions:
\textit{camera}, \textit{robot}, \textit{light}, and \textit{layout}. These
dimensions are chosen because they are closely related to spatial perception
and provide a suitable test suite for assessing how implicit depth cues affect
robustness under viewpoint, embodiment, illumination, and scene arrangement
variations.

\noindent\textbf{Progressive Alignment Training Ablation.}
To evaluate the proposed Progressive Alignment Training strategy, we compare
one-stage, two-stage, and three-stage training variants on the LIBERO Long
benchmark. In the one-stage variant, all modules are jointly optimized from
the beginning. In the two-stage variant, Stage 1 trains only SEM and the
Action Expert while IDEM and the Vision-Language Backbone are frozen, and
Stage 2 unfreezes the entire model for end-to-end optimization. In the
three-stage variant, Stage 1 also trains only SEM and the Action Expert, Stage
2 additionally unfreezes IDEM while keeping the Vision-Language Backbone
frozen, and Stage 3 unfreezes the entire model for end-to-end optimization.
This ablation examines whether more gradual cross-module alignment improves
long-horizon manipulation performance.

\noindent\textbf{Fusion Strategy Ablation.}
To investigate the impact of different fusion strategies, we compare three
ways of integrating the implicit depth features extracted by IDEM into the 2D
vision-language representations. \textit{Concat} directly concatenates the
depth features with the 2D representations before the Action Expert.
\textit{Crossatten} uses the 2D representations as queries and the depth
features as keys and values for cross-attention fusion. \textit{SEM} first
aligns the depth features to the hidden space of the 2D representations and
then injects them through FiLM-style feature-wise modulation. This ablation
examines how different fusion mechanisms affect the use of implicit depth cues
for action prediction.

\section{Implementation Details and Reproducibility}

Unless otherwise specified, Evo-Depth is trained with the three-stage
progressive alignment strategy described in Section~3. In Stage 1, only the
SEM and Action Expert are optimized, while the vision-language backbone and
IDEM remain frozen. In Stage 2, IDEM is unfrozen and jointly optimized with
the SEM and Action Expert, while the vision-language backbone remains frozen.
In Stage 3, all modules are jointly optimized end-to-end. For sequential
training, Stage 2 is initialized from the final checkpoint of Stage 1, and
Stage 3 is initialized from the final checkpoint of Stage 2.

Table~\ref{tab:impl_metaworld} lists a representative set of training
hyperparameters for the Meta-World setting. The remaining simulation
benchmarks follow the same overall training recipe unless otherwise specified,
with benchmark-dependent dataset configurations and stage lengths. Across all
stages, we use AdamW with a linear warmup followed by cosine learning-rate
decay.

\begin{table}[t]
\centering
\small
\caption{\textbf{Representative Training Hyperparameters on Meta-World.}}
\label{tab:impl_metaworld}
\setlength{\tabcolsep}{3.5pt}
\renewcommand{\arraystretch}{0.9}
\begin{tabular*}{\columnwidth}{@{}p{4.8cm}@{\extracolsep{\fill}}l@{}}
\toprule
Setting & Value \\
\midrule
Vision-language backbone & InternVL3-1B \\
Action head & Flow matching \\
Input image size & 448 \\
Data augmentation & Enabled \\
Batch size & 16 \\
Optimizer & AdamW \\
Learning rate & $1\times10^{-5}$ \\
Learning-rate schedule & Linear warmup + cosine decay \\
Warmup steps & 1{,}000 \\
Weight decay & $1\times10^{-3}$ \\
Gradient clipping norm & 1.0 \\
Action horizon & 50 \\
Action dimension & 24 \\
State dimension & 24 \\
Action head layers & 8 \\
Dropout & 0.2 \\
Data-loading workers & 4 \\
Stage 1 steps & 5{,}000 \\
Stage 2 steps & 10{,}000 \\
Stage 3 steps & 120{,}000 \\
Stage 1 trainable modules & SEM + Action Expert \\
Stage 2 trainable modules & IDEM + SEM + Action Expert \\
Stage 3 trainable modules & VLM + IDEM + SEM + Action Expert \\
\bottomrule
\end{tabular*}
\renewcommand{\arraystretch}{1}
\end{table}

\section{Compute Resources}
Unless otherwise specified, training for both simulation and real-world
settings was conducted on NVIDIA A100 GPUs. As a representative example,
Meta-World training used 8 GPUs under the three-stage training pipeline
described above. The remaining training runs follow the same overall hardware
setting, with benchmark-dependent variation in dataset scale and training
length.

For real-robot deployment, inference efficiency is measured on an NVIDIA RTX
4090D platform, as reported in the main paper.

\section{Assets and Licenses}
We build on several publicly available models and benchmarks. We cite the
corresponding original papers in the main text and use these assets under
their stated licenses or benchmark terms. The main external assets used in
this work are summarized in Table~\ref{tab:assets_licenses}.

\begin{table}[t]
    \centering
    \small
    \caption{\textbf{External assets and licenses.}}
    \label{tab:assets_licenses}
    \setlength{\tabcolsep}{3pt}
    \renewcommand{\arraystretch}{0.9}
    \begin{tabular*}{\columnwidth}{@{\extracolsep{\fill}}p{5.0cm}p{2.2cm}}
        \toprule
        Asset & License / terms \\
        \midrule
        InternVL3-1B & Apache-2.0 \\
        Depth Anything 3 (DA3-BASE) & Apache-2.0 \\
        Meta-World & MIT \\
        VLA-Arena & Apache-2.0 \\
        LIBERO & MIT \\
        LIBERO-Plus & MIT \\
        \bottomrule
    \end{tabular*}
    \renewcommand{\arraystretch}{1}
\end{table}

\section{Limitations}
Our current real-world evaluation focuses on representative tabletop
manipulation tasks. While these tasks cover object grasping, placement, and
stacking scenarios with different spatial demands, future work can further
broaden the evaluation to a wider range of manipulation tasks and scene
configurations.

In addition, this study focuses on fixed-base manipulation rather than broader
embodied settings such as mobile manipulation or whole-body interaction. We
view this as a scope choice consistent with the goal of studying lightweight
depth-enhanced spatial perception for manipulation, and extending the
framework to more general embodied platforms is an interesting direction for
future work.

\section{Broader Impacts}
This work may contribute to more efficient and practical robotic manipulation
by improving spatially grounded policy learning under lightweight RGB-only
settings. In the long term, such systems may benefit application scenarios
such as laboratory automation, industrial assistance, and other structured
manipulation environments where reliable and efficient robot operation is
important.

At the same time, if such systems are deployed without sufficient safeguards,
errors in perception or action prediction may lead to failed grasps or object
damage. We therefore believe that real-world deployment should still be
accompanied by appropriate safety constraints and human oversight.

\end{document}